\documentclass[letterpaper]{article} 
\usepackage[preprint]{aaai2027}  
\usepackage[hyphens]{url}  
\usepackage{graphicx} 
\usepackage{natbib}  
\usepackage{caption} 
\usepackage{algorithm}
\usepackage{algorithmic}
\usepackage{newfloat}
\usepackage{listings}
\DeclareCaptionStyle{ruled}{labelfont=normalfont,labelsep=colon,strut=off} 
\floatstyle{ruled}
\newfloat{listing}{tb}{lst}{}
\floatname{listing}{Listing}
\usepackage{booktabs}
\usepackage{amsmath}
\usepackage{amssymb}

\title{SafeRI: Recognition and Intervention for Token-Level Safety Intervention in Large Vision Language Models}

\author{
    Caoyuan Ma\textsuperscript{\rm 2,\rm 3}\equalcontrib,
    Tian Gu\textsuperscript{\rm 1,\rm 5}\equalcontrib,
    Wenpu Liu\textsuperscript{\rm 3,\rm 4},
    Weichu Xie\textsuperscript{\rm 3,\rm 4},
    Shuai Dong\textsuperscript{\rm 3,\rm 6},\\
    Yuqi Xu\textsuperscript{\rm 3,\rm 4},
    Ji Zhao\textsuperscript{\rm 3},
    Ziyue Wang\textsuperscript{\rm 3,\rm 4},
    Wenzheng Chang\textsuperscript{\rm 3,\rm 7},
    Taiqiang Wu\textsuperscript{\rm 3,\rm 8},\\
    Yongfu Zhu\textsuperscript{\rm 3},
    Wenqi Shao\textsuperscript{\rm 3}\thanks{Project leader.},
    Zheng Wang\textsuperscript{\rm 1}\thanks{Co-corresponding authors.},
    Yinqiang Zheng\textsuperscript{\rm 2}\footnotemark[3]
}
\affiliations{
    \textsuperscript{\rm 1}Wuhan University\quad
    \textsuperscript{\rm 2}The University of Tokyo\quad
    \textsuperscript{\rm 3}JD.com\quad
    \textsuperscript{\rm 4}Peking University\\
    \textsuperscript{\rm 5}Shanghai AI Laboratory\quad
    \textsuperscript{\rm 6}Shanghai Innovation Institute\quad
    \textsuperscript{\rm 7}Shanghai Jiao Tong University\quad
    \textsuperscript{\rm 8}The University of Hong Kong
}

\begin{document}

\maketitle

\begin{abstract}
Existing safety alignment methods for vision-language models usually modify the model behavior globally: once the safety parameters are trained or loaded, they participate in both unsafe and already-safe generations. This always-on intervention can unnecessarily perturb the model's original reasoning path and degrade general multimodal capabilities. We argue that safety alignment should be an on-demand intervention rather than a permanent modification to every decoding trajectory. To this end, we propose a streaming recognition and gated LoRA framework for intrinsic VLM safety. During autoregressive generation, a lightweight recognizer estimates whether the current pre-token generation state is safe or unsafe. Its output updates the LoRA gate for the following decoding step; otherwise, generation follows the frozen-backbone policy. The LoRA module is trained from unsafe prefixes, transition statements, and safe continuations, so that it learns to redirect unsafe generations back to safe responses after activation. Experiments across multiple safety and general-purpose benchmarks demonstrate the effectiveness of our method in post-alignment settings.
\end{abstract}

\begin{links}
    \link{Project page}{https://safe-vlm.github.io/SafeRI/}
\end{links}

\section{Introduction}
\label{sec:intro}

Safety risk in vision-language models (VLMs) is not determined by the input alone; it materializes along the model's autoregressive generation trajectory. The same harmful-looking multimodal request may be safely refused by one decoding trajectory but answered with actionable details by another, while an initially ambiguous request may reveal its risk only as the response unfolds. This asymmetry is especially important for VLMs, where harmful intent can be distributed across images, text, and their interaction \citep{ye2025survey,qi2023visualadversarial,gong2023figstep,shayegani2024jailbreakpieces}. It is even more pronounced when hardening an instruction-tuned VLM that has already undergone safety alignment: such a model safely handles most benign requests and may already refuse many harmful ones, leaving residual failures concentrated in a relatively small subset of generation trajectories. In this regime, whether additional intervention is needed---and when it becomes necessary---can only be determined from the evolving generation state rather than from a one-shot judgment of the prompt.

Existing VLM safety methods do not fully match this temporal nature of risk. Input-level defenses must decide before observing what the model will generate, whereas output-level guards act only after a potentially unsafe response has already been produced. Training-time alignment and always-on safety adapters \citep{zong2024vlguard,zhang2024spavl,liu2024safetyalignment,wang2024adashield,oh2024uniguard,xu2024cider} avoid this timing problem by continuously biasing the model toward safe behavior, but they also modify benign trajectories and harmful requests that the base model would already refuse. Such unnecessary intervention creates a \emph{safety-alignment tax}: perturbed hidden states and token probabilities can lead to over-refusal, stylistic drift, or degraded multimodal utility even when no correction is needed.

\begin{figure}[t]
\centering
\includegraphics[width=\columnwidth]{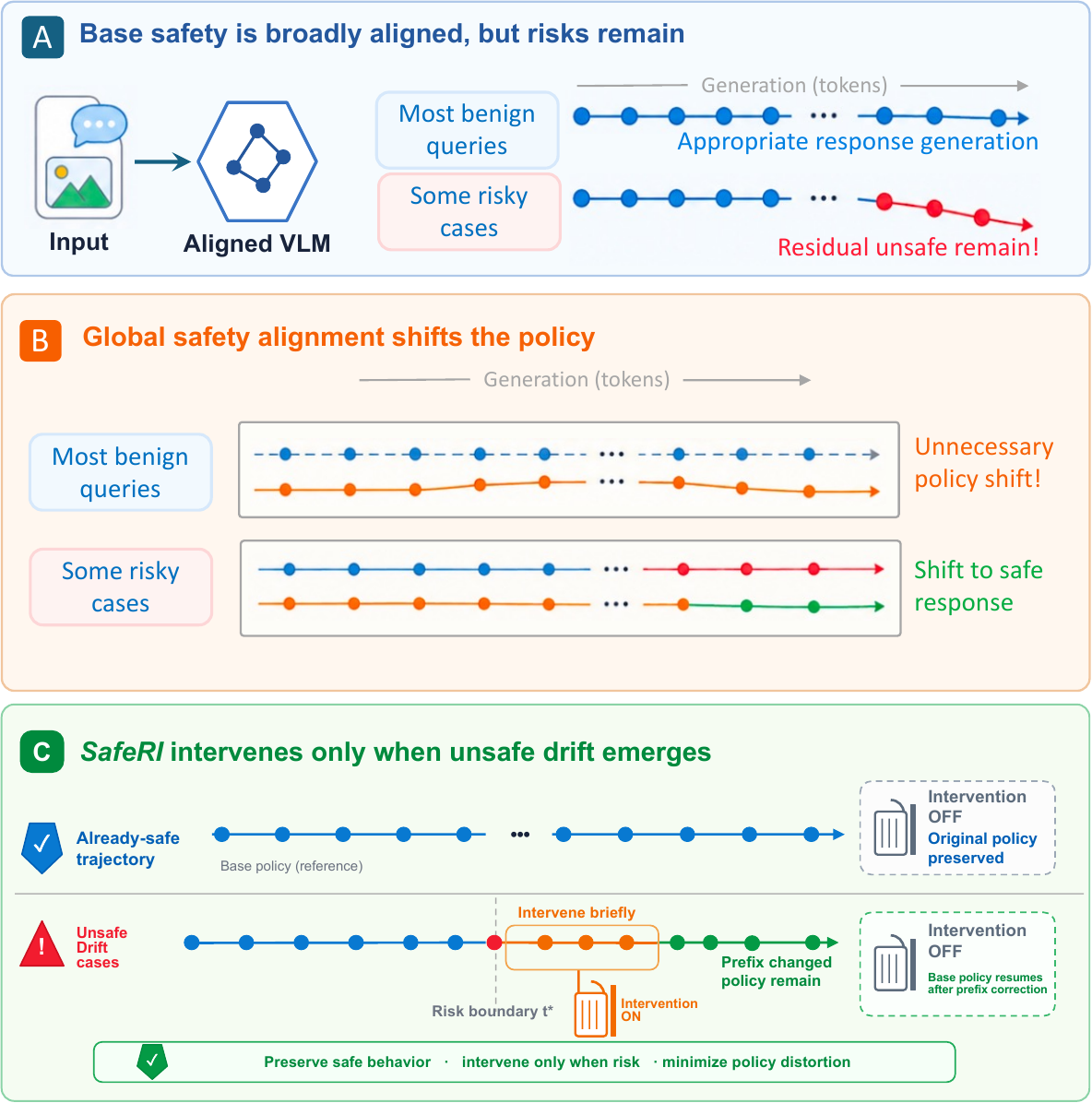}
\caption{Comparison between always-on safety fine-tuning and SafeRI. Global safety alignment can unnecessarily perturb safe decoding trajectories, while SafeRI activates safety parameters only when unsafe generation states emerge, preserving the frozen VLM path for safe responses.}
\label{fig:method-overview}
\end{figure}

These limitations suggest a different formulation: \textbf{post-alignment safety hardening should be an on-demand trajectory-control problem, rather than a global relearning of safety behavior.} The system must continuously recognize whether the partial response is approaching unsafe content and, once risk emerges, modify the token-generation policy for the remaining tokens. These two operations must be coupled in time. Recognition without immediate intervention only reports a failure, while intervention without trajectory-level recognition either acts too early or perturbs every response. Conversely, as long as the frozen backbone remains on a safe trajectory, the appropriate action is to preserve the safety behavior and general capabilities it already possesses.

We instantiate this formulation as SafeRI, a \emph{Recognition and Intervention} framework for token-level safety intervention. At each decoding step, a lightweight streaming detector reads the evolving hidden state and predicts whether generating the next token is likely to move the trajectory into an unsafe region. When no such imminent risk is predicted, the gate remains closed and generation continues under the frozen-backbone policy. Otherwise, the gate activates a parameter-efficient LoRA behavior-correction module \citep{hu2022lora} for the next and subsequent tokens, injecting a localized update that redirects the response toward a safe continuation. Unlike input-conditioned routing, which selects a model or module before decoding, SafeRI repeatedly uses the partial assistant trajectory to decide whether the safety adapter should modify the policy for subsequent tokens. Recognition thus decides \emph{when} to intervene, while intervention determines \emph{how} the future trajectory is changed.

We train the two components with supervision aligned to the onset of unsafe generation. The recognizer learns whether the current prefix predicts an unsafe next step, while LoRA learns from risky prefixes, transition statements, and safe continuations to redirect subsequent generation. This shared boundary makes the gate actionable while preserving the distinct roles of recognition and intervention.

Our contributions are as follows:
\begin{itemize}
    \item We formulate post-alignment VLM safety hardening as a selective trajectory-control problem: the safety adapter should be invoked according to the model's evolving response, while already-safe trajectories should retain the frozen-backbone policy.
    \item We propose SafeRI, which detects emerging unsafe trajectories from prefix hidden states during decoding and activates a boundary-aligned LoRA trained on risky prefixes and safe continuations to redirect subsequent generation. Once risk subsides, the gate closes and generation resumes under the frozen-backbone policy.
    \item Experiments across multiple VLM families show that SafeRI mitigates residual risk while preserving general multimodal capability, with ablations characterizing the effects of gate thresholds, adapter placement, and activation causes.
\end{itemize}

\begin{figure*}[t]
\centering
\includegraphics[width=\textwidth]{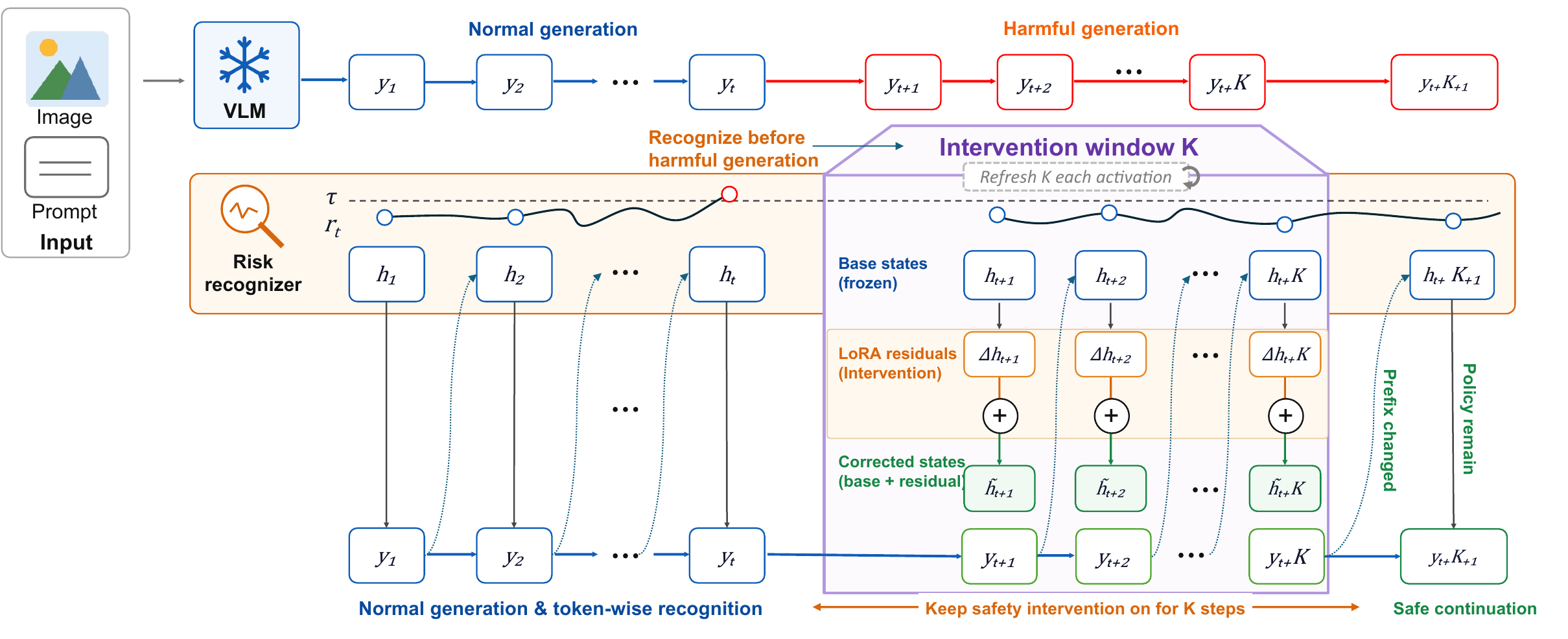}
\caption{SafeRI at inference time. $y_t$ denotes the final token of the pre-risk prefix. After $y_t$ is appended, the risk recognizer scores the current-prefix state $h_t$. If $r_t>\tau$, the safety LoRA is activated beginning with the generation of $y_{t+1}$ for a renewable $K$-step window; each new threshold crossing resets the window to $K$. While active, LoRA residuals $\Delta h$ redirect the frozen backbone toward a safe continuation. The upper and lower trajectories show the uncorrected harmful generation and the corrected generation, respectively.}
\label{fig:method-overview-diagram}
\end{figure*}

\section{Related Work}
\label{sec:related}


Prior safety methods differ in \emph{when} they intervene and \emph{how} they alter model behavior. Training-time alignment changes the deployed model globally, inference-time guardrails act on inputs, outputs, or decoding, and internal interventions modify representations or token distributions. SafeRI instead asks when additional safety parameters should be activated as a response evolves.

\paragraph{Training-Time Safety Alignment.}
Training-time defenses improve intrinsic safety through fine-tuning, preference alignment, and safety-oriented data. VLGuard studies targeted safety fine-tuning \citep{zong2024vlguard}, SPA-VL supplies large-scale multimodal preference data \citep{zhang2024spavl}, and related work develops alignment recipes and data-curation pipelines \citep{liu2024safetyalignment,helff2024llavaguard}. Other methods refine the safety signal through representation constraints, decoupled refusal training, token-level data selection, or constraints on safety-critical tokens \citep{chen2025learning,yuan2025refuse,li2026token,wang2026few}; safety-oriented reasoning models further introduce verifier-guided reinforcement learning and constructive alignment \citep{lab2025safeworkr1coevolvingsafetyintelligence,duan2025oyster}. Despite these differences, they encode safety into a fixed deployed parameter set. SafeRI instead keeps the backbone frozen and activates learned safety parameters only when the evolving generation indicates risk.

\paragraph{Inference-Time Guardrails.}
Inference-time defenses avoid updating the full backbone. They include prompt-based guardrails \citep{wang2024adashield,oh2024uniguard}, detectors for inconsistent or perturbed multimodal inputs \citep{xu2024cider,zhang2023jailguard}, auxiliary protection or modality-transformation pipelines \citep{pi2024mllmprotector,gou2024ecso}, and guard models that classify inputs or responses \citep{inan2023llamaguard}. More closely related, Safety Reminder reactivates safety awareness with a soft prompt, while CASA uses conditional decoding for multimodal safety \citep{tang2025safetyremindersoftprompt,kumar2026robust}. SafeRI instead monitors the partial answer and activates learned safety parameters only when the trajectory becomes risky, avoiding unnecessary intervention when the backbone is already refusing safely.

\paragraph{Internal Intervention and Conditional Adaptation.}
Internal-intervention methods directly steer activations, predictions, or decoding objectives at inference time \citep{wang2024inferaligner,gao2024coca,ghosal2024immune}. Separately, LoRA enables parameter-efficient adaptation with a frozen backbone \citep{hu2022lora}; LoRA-Guard applies it to content moderation, and Activated LoRA (aLoRA) supports conditional adapter invocation \citep{elesedy2024loraguard,greenewald2025activated}. Neither internal steering nor conditional adaptation alone determines whether an evolving response requires correction. SafeRI connects them through hidden-state risk recognition: its safety LoRA remains dormant on safe trajectories and activates token by token when risk emerges.

\section{Method}
\label{sec:method}

\subsection{Problem Formulation}

Let $q$ denote the complete multimodal user input. A VLM generates a response autoregressively:
\begin{equation}
    y_t \sim p_\theta(y_t \mid q, y_{<t}).
\end{equation}
Conventional safety tuning changes the model distribution globally. SafeRI
instead couples Recognition and Intervention across consecutive decoding
steps. Before token $y_t$ is sampled, the gate $g_t$, carried over from the
previous step, configures the forward pass on the current prefix $y_{<t}$.
This forward pass produces both the logits for $y_t$ and a pre-token hidden
state inspected by the recognizer. The resulting risk score updates the gate
$g_{t+1}$ for the following forward pass; it does not retroactively change the
logits already computed for $y_t$. The distribution used to sample the current
token is therefore
\begin{equation}
    p(y_t \mid q, y_{<t}) =
    \begin{cases}
    p_\theta(y_t \mid q, y_{<t}), & g_t = 0, \\
    p_{\theta, \phi}(y_t \mid q, y_{<t}), & g_t = 1,
    \end{cases}
\end{equation}
where $\theta$ denotes the frozen VLM backbone, $\phi$ denotes the safety
LoRA parameters, and $g_t$ controls Intervention during the forward pass that
produces the logits for $y_t$. The objective is to keep the gate closed unless
a preceding recognition result indicates that intervention is needed.

\subsection{Streaming Risk Recognition}

At decoding step $t$, the model runs a forward pass on the current prefix
$y_{<t}$ under gate $g_t$. Let $H_t$ denote the sequence of final-layer
hidden states from this pre-token forward pass and let $\mathbf{z}_t$ denote
the next-token logits at its last position:
\begin{equation}
    \begin{aligned}
        (H_t,\mathbf{z}_t) &= f_{\theta,\phi}^{(g_t)}(q,y_{<t}), \\
        H_t &\in\mathbb{R}^{B\times L_t\times d}, \\
        \mathbf{z}_t &\in\mathbb{R}^{B\times|\mathcal{V}|}.
    \end{aligned}
\end{equation}
where $f_{\theta,\phi}^{(g_t)}$ denotes the current forward pass with the
LoRA disabled or enabled according to $g_t$, $B$ is the batch size, $L_t$ is
the current sequence length, and $d$ is the backbone hidden size. The
recognizer is not inserted into any transformer layer. It is a shared linear
head applied after the backbone's final layer:
\begin{equation}
    R_\psi(H_t)=H_tW_\psi+b_\psi
    \in\mathbb{R}^{B\times L_t\times 2},
\end{equation}
where $W_\psi\in\mathbb{R}^{d\times 2}$,
$b_\psi\in\mathbb{R}^{2}$, and class indices $0$ and $1$ correspond to
\emph{Safe} and \emph{Unsafe}, respectively. During inference, only the last
prefix position is used:
\begin{equation}
    h_t=H_t[:,-1,:], \qquad
    r_t=\left[\operatorname{softmax}(R_\psi(h_t))\right]_1.
\end{equation}
Thus, $r_t$ is the Unsafe probability assigned to the current prefix's last
hidden state. Although it is computed before sampling, the logits
$\mathbf{z}_t$ have already been produced by the same forward pass. Therefore,
$r_t$ updates Intervention for the next forward pass rather than changing the
distribution from which $y_t$ is sampled:
\begin{equation}
    y_t\sim\operatorname{softmax}(\mathbf{z}_t), \qquad
    r_t \longrightarrow g_{t+1}.
\end{equation}

A threshold crossing updates a renewable sticky counter for the following
decoding step:
\begin{equation}
    \begin{aligned}
        c_{t+1} & =
        \begin{cases}
            K, & r_t>\tau, \\
            \max(c_t-1,0), & r_t\leq\tau,
        \end{cases} \\
        g_{t+1} & =\mathbb{I}[c_{t+1}>0].
    \end{aligned}
\end{equation}
where $\tau$ is the chosen operating threshold, $K$ is the sticky duration,
and $c_t$ records the activation budget available to the current step. A
threshold crossing at $h_t$ sets $c_{t+1}=K$ and first affects the forward
pass that produces the logits for $y_{t+1}$. The window is renewable: if the
detector again exceeds the threshold while the LoRA is active, the next-step
counter is reset to $K$ rather than continuing to expire. Consequently, the
gate closes only after the renewed $K$-step activation horizon expires without
another threshold crossing. This mechanism sustains intervention while risk
persists but allows the model to return automatically to the frozen-backbone
path once the trajectory remains stable. It does not alter $\mathbf{z}_t$ or $y_t$.

\subsection{Gated LoRA Intervention}

\begin{algorithm}[t]
\caption{Streaming Gated-LoRA Decoding}
\begin{algorithmic}[1]
\STATE Input: multimodal user input $q$, frozen VLM $\theta$, LoRA $\phi$, recognizer $R_\psi$
\STATE Initialize $y_{<1}=\emptyset$, gate $g_1=0$, and sticky counter $c_1=0$
\FOR{$t=1$ to $T$}
    \STATE Compute $(H_t,\mathbf{z}_t)=f_{\theta,\phi}^{(g_t)}(q,y_{<t})$
    \STATE Set $h_t=H_t[:,-1,:]$ and $r_t=[\operatorname{softmax}(R_\psi(h_t))]_1$
    \STATE Sample $y_t$ from $\operatorname{softmax}(\mathbf{z}_t)$ and append it to the prefix
    \IF{$r_t > \tau$}
        \STATE $c_{t+1} \leftarrow K$
    \ELSE
        \STATE $c_{t+1}\leftarrow\max(c_t-1,0)$
    \ENDIF
    \STATE Set $g_{t+1}\leftarrow\mathbb{I}[c_{t+1}>0]$
\ENDFOR
\end{algorithmic}
\end{algorithm}

The LoRA module is inserted into selected transformer layers. For a target linear projection $W_l$ in layer $l$, the gated projection is:
\begin{equation}
    W_l^{(g_t)} = W_l + g_t\frac{\alpha}{r}B_lA_l,
\end{equation}
where $A_l$ and $B_l$ are rank-$r$ LoRA factors and $\alpha$ is the LoRA scaling coefficient. Thus, $g_t=0$ disables every selected adapter projection for the forward pass that produces $H_t$ and $\mathbf{z}_t$, whereas $g_t=1$ enables the adapter update in those projections to redirect the distribution used to sample $y_t$ and the subsequent continuation.

This design creates three expected behaviors:
\begin{itemize}
    \item For benign multimodal tasks, the detector is expected to remain inactive; for every token produced while it is inactive, the next-token policy is exactly that of the frozen backbone.
    \item For harmful prompts that the base model already refuses, the detector can remain inactive because the assistant trajectory does not approach an unsafe onset.
    \item When the current pre-token state is judged risky, the detector opens the gate for the following forward pass and the subsequent sticky-window steps.
\end{itemize}

\subsection{Training Data Construction}

Each training instance contains four components:
\begin{equation}
    (q, u^{-}, b, s),
\end{equation}
where $q$ is the multimodal user input, $u^{-}$ is the assistant prefix immediately before the first clearly risky token in an unsafe trajectory, $b$ is a short transition statement, and $s$ is a safe continuation. The original risky suffix beginning at that onset is discarded, and the rewritten target is constructed as:
\begin{equation}
    u^{-} \; + \; b \; + \; s,
\end{equation}
where $b$ is a transition statement such as ``I should stop here'' or ``I cannot continue with these instructions.'' The transition statement helps the model learn a natural local correction from unsafe intent to safe refusal or safe redirection.

The LoRA is trained to produce the transition and safe continuation after
observing the pre-risk prefix. The recognizer is supervised on the pre-token
state associated with this boundary. In streaming decoding, the score from
that state updates the gate for the following forward pass because the current
logits have already been computed. The intervention is therefore causal but
not retroactive. Importantly, the goal is not to memorize complete refusal
templates, but to redirect the continuation toward a safe local correction
once the gate becomes active.
The complete data construction and cleaning pipeline is described in the
appendix.

\subsection{Two-Stage Training Objective}

We train the detector and the gated LoRA in two decoupled stages while keeping the base VLM frozen. The first stage learns \emph{when} a trajectory is about to enter an unsafe region, and the second stage learns \emph{how} to redirect the subsequent generation. This separation prevents the classifier from being optimized for text generation and prevents the LoRA module from learning the trigger boundary implicitly.

\paragraph{Stage 1: risk-boundary classifier.}
In the first stage, only the risk-boundary classifier $D_\psi$ is trained. For
an assistant trajectory $y$, the frozen backbone computes each pre-token state
$h_t=f_\theta(q,y_{<t})$. Let $t_b$ denote the manually specified boundary
index such that $y_{<t_b}=u^{-}$. The classifier predicts a binary distribution
over ordinary and imminent-risk states:
\begin{equation}
    \begin{aligned}
    \mathbf{p}_\psi(\cdot \mid h_t)
        &= \operatorname{softmax}(D_\psi(h_t)), \\
    a_t
        &\in \{\text{Ordinary},\text{ImminentRisk}\}.
    \end{aligned}
\end{equation}
The supervision is attached to pre-token prefix states rather than to the
safety category of an already sampled token. Sampled prompt states and early
pre-risk-prefix states are labeled Ordinary, while the boundary state
$h_{t_b}=f_\theta(q,u^{-})$ is labeled ImminentRisk. In our default setting,
only this single boundary state receives the positive label. Positions that do
not participate in classifier supervision are masked out.

\paragraph{Boundary-position trade-off.}
The boundary position is specified during offline data construction, which
provides direct supervision for the exact state at which the gate should first
request intervention for the following decoding step. This explicit alignment
makes the detector lightweight, matches the one-step actuation delay in
streaming decoding, and avoids labeling an entire harmful-looking prompt or
response as unsafe when only its continuation requires correction. However,
the onset of unsafe content is not always a single unambiguous token: it may
emerge gradually, depend on context, or shift under tokenization. The manually
specified boundary can therefore introduce annotation noise and sensitivity to
the chosen boundary policy. Placing it too early favors safety recall but may
cause unnecessary intervention and utility loss, whereas placing it too late
preserves more of the original trajectory but may allow unsafe content to
appear before the gate activates. The boundary policy must therefore account
for the one-step delay between recognition and intervention. Our selected
boundary is a practical causal target rather than a claim that unsafe onset has
a uniquely defined position.

Let $a_t\in\{0,1\}$ denote the binary label attached to pre-token state
$h_t$, where $0$ is Ordinary and $1$ is ImminentRisk, and let $m_t$ indicate
whether state position $t$ is supervised. The Stage 1 objective is masked cross
entropy:
\begin{equation}
    \begin{aligned}
    \mathcal{L}_{\text{stage1}}
        &= -\frac{1}{|\mathcal{M}|}
        \sum_{t\in\mathcal{M}} \log p_\psi(a_t \mid h_t), \\
    \mathcal{M}
        &= \{t \mid m_t=1\}.
    \end{aligned}
\end{equation}
During this stage, the base model parameters $\theta$ and the LoRA parameters $\phi$ are frozen; only the classifier head parameters $\psi$ are updated. Thus, Stage 1 learns a token-level gate signal without changing the generation behavior of the VLM.

\paragraph{Stage 2: LoRA safety rewrite.}
In the second stage, the classifier and base VLM remain frozen, and only the LoRA parameters $\phi$ are trained. Each rewritten answer is decomposed into a pre-risk prefix $u^{-}$, a transition statement $b$, and a safe continuation $s$. The offline boundary markers used to construct these segments are removed before training, so the model input is the concatenation of the prompt and $u^{-}+b+s$.

Stage 2 uses causal language modeling supervision on both the transition statement and the safe continuation. Tokens belonging to the prompt and pre-risk prefix are used as context but assigned the ignore label. Formally, for token position $t$,
\begin{equation}
    \ell_t =
    \begin{cases}
    y_t, & y_t \in b \oplus s, \\
    \texttt{ignore}, & \text{otherwise}.
    \end{cases}
\end{equation}
With LoRA enabled in the selected layers, the model produces logits from $f_{\theta,\phi}$. The Stage 2 objective is the masked autoregressive language modeling loss:
\begin{equation}
    \mathcal{L}_{\text{stage2}}
    = -\frac{1}{|\mathcal{S}_{b+s}|}
    \sum_{t\in\mathcal{S}_{b+s} }
    \log p_{\theta,\phi}(y_t \mid q, y_{<t}),
\end{equation}
where $\mathcal{S}_{b+s}$ is the set of transition-and-safe-continuation token positions. In implementation, this corresponds to the standard shifted causal language modeling loss with all prompt and prefix labels set to an ignore index. If truncation removes all supervised $b\oplus s$ tokens, the sample is skipped for Stage 2 updates.

\paragraph{Overall optimization and inference coupling.}
The complete training procedure is therefore a staged optimization:
\begin{equation}
    \psi^* = \arg\min_\psi \mathcal{L}_{\text{stage1}}(\psi;
    \theta \text{ frozen}, \phi \text{ frozen}),
\end{equation}
\begin{equation}
    \phi^* = \arg\min_\phi \mathcal{L}_{\text{stage2}}(\phi;
    \theta \text{ frozen}, \psi^* \text{ frozen}).
\end{equation}
At inference time, gate $g_t$ first configures the forward pass on
$(q,y_{<t})$, which jointly produces the final-layer state sequence $H_t$ and
next-token logits $\mathbf{z}_t$. The recognizer evaluates the last-position
state $h_t=H_t[:,-1,:]$ before $y_t$ is sampled. If
$r_t$ exceeds $\tau$, it sets $g_{t+1}=1$ through the sticky-counter update,
so the trained LoRA is first applied to the following forward pass. When the
gate is closed, the selected adapter projections are disabled and that forward
pass follows the frozen-backbone policy.

\section{Experiments}
\label{sec:experiments}

\begin{table*}[t]
\centering
\normalsize
\begin{tabular*}{\textwidth}{@{\extracolsep{\fill}}llcccccc@{}}
\toprule
Model & Setting & SPAVL-test harm & AdvBench & HADES & XSTest & MSSBench & Safety Avg. \\
\midrule
Qwen3.5-9B & Base & 95.47 & 99.62 & 94.89 & 90.50 & 52.33 & 86.56 \\
& SafeRI & 99.25 & 99.42 & 98.31 & 90.50 & 56.92 & 88.88 \\
& Always-on & 100.00 & 99.62 & 100.00 & 81.33 & 66.12 & 89.41 \\
& DPO & 100.00 & 99.62 & 99.91 & 86.89 & 57.81 & 88.85 \\
\midrule
Qwen3.5-4B & Base & 96.60 & 99.62 & 97.91 & 90.22 & 53.52 & 87.57 \\
& SafeRI & 99.25 & 99.81 & 99.11 & 89.00 & 54.25 & 88.28 \\
& Always-on & 100.00 & 99.81 & 100.00 & 79.78 & 59.54 & 87.83 \\
& DPO & 100.00 & 90.96 & 100.00 & 77.11 & 52.76 & 84.17 \\
\midrule
Qwen3.5-2B & Base & 96.98 & 98.85 & 96.40 & 87.78 & 52.70 & 86.54 \\
& SafeRI & 98.11 & 98.85 & 97.96 & 89.50 & 54.08 & 87.70 \\
& Always-on & 100.00 & 100.00 & 100.00 & 69.33 & 54.08 & 84.68 \\
& DPO & 99.87 & 90.96 & 99.96 & 74.67 & 50.46 & 83.18 \\
\midrule
Llama3.2-Vision-11B & Base & 89.06 & 90.50 & 95.87 & 89.56 & 52.45 & 83.49 \\
& SafeRI & 92.08 & 93.27 & 95.68 & 85.50 & 54.84 & 84.27 \\
& Always-on & 95.47 & 94.40 & 98.00 & 81.10 & 59.25 & 85.64 \\
& DPO & 97.82 & 98.24 & 97.82 & 84.96 & 54.84 & 86.74 \\
\midrule
SafeWork-R1 & -- & 97.36 & 95.38 & 97.91 & 92.22 & 67.70 & 90.11 \\
SafeProbing & -- & 98.87 & 99.42 & 99.73 & 88.22 & 54.09 & 88.07 \\
DTR & -- & 97.36 & 99.42 & 96.40 & 87.56 & 49.95 & 86.14 \\
IMMUNE & -- & 97.74 & 77.88 & 97.82 & 89.11 & 49.04 & 82.32 \\
\bottomrule
\end{tabular*}
\caption{Main safety evaluation. Higher scores indicate safer behavior. Safety Avg. is the arithmetic mean of the five benchmarks. SafeWork-R1 is an external 72B checkpoint; SafeProbing and DTR are controlled Qwen3.5-9B reproductions, and IMMUNE is a controlled adaptation to the same backbone.}
\label{tab:safety-main}
\end{table*}
\begin{table}[t]
\centering
\normalsize
\setlength{\tabcolsep}{2pt}
\begin{tabular}{@{}llcccc@{}}
\toprule
Model & Setting & \shortstack{MM\\Bench} & MM-Vet & BLINK & \shortstack{General\\Avg.} \\
\midrule
Qwen3.5-9B & Base & 69.44 & 79.29 & 54.81 & 67.85 \\
& SafeRI & 68.04 & 78.40 & 56.49 & 67.64 \\
& Always-on & 62.28 & 64.06 & 54.97 & 60.44 \\
& DPO & 44.21 & 49.29 & 41.35 & 44.95 \\
\midrule
Qwen3.5-4B & Base & 65.86 & 79.45 & 53.66 & 66.32 \\
& SafeRI & 63.83 & 77.18 & 53.55 & 64.85 \\
& Always-on & 49.55 & 66.72 & 55.23 & 57.17 \\
& DPO & 38.14 & 53.70 & 53.79 & 48.54 \\
\midrule
Qwen3.5-2B & Base & 64.36 & 70.76 & 48.03 & 61.05 \\
& SafeRI & 73.37 & 66.14 & 47.13 & 62.21 \\
& Always-on & 32.78 & 53.65 & 48.03 & 44.82 \\
& DPO & 33.26 & 44.62 & 43.01 & 40.30 \\
\midrule
\shortstack[l]{Llama3.2-\\Vision-11B} & Base & 64.05 & 67.80 & 46.50 & 59.45 \\
& SafeRI & 62.70 & 64.90 & 49.20 & 58.93 \\
& Always-on & 60.11 & 59.28 & 45.12 & 54.84 \\
& DPO & 49.76 & 47.43 & 40.24 & 45.81 \\
\midrule
SafeWork-R1 & -- & 61.43 & 69.50 & 64.07 & 65.00 \\
SafeProbing & -- & 56.25 & 61.57 & 53.87 & 57.23 \\
DTR & -- & 83.00 & 60.07 & 55.26 & 66.11 \\
IMMUNE & -- & 74.00 & 60.14 & 51.57 & 61.90 \\
\bottomrule
\end{tabular}
\caption{General multimodal capability. Higher scores are better.}
\label{tab:general-mm}
\end{table}

\subsection{Experimental Setup}

\paragraph{Model and Data}
We use Qwen3.5-9B as the main backbone and additionally evaluate Qwen3.5-2B, Qwen3.5-4B, and Llama3.2-Vision-11B to test transfer across model scales and families. All training supervision is constructed by relabeling examples from the publicly released SPA-VL dataset, without introducing additional training data.

\paragraph{Benchmarks} We evaluate five safety benchmarks and three general multimodal benchmarks using deterministic decoding with temperature $0$. Evaluations requiring an LLM judge use \texttt{gpt-oss-20b}. Benchmark protocols, manual-verification procedures, and multi-seed results are provided in the appendix.

\paragraph{Baseline Protocols} We report SafeWork-R1 as an external 72B-checkpoint reference. For controlled comparisons, we reproduce SafeProbing and DTR and adapt IMMUNE to the Qwen3.5-9B backbone, evaluating all three under the same decoding and benchmark settings as SafeRI.


We evaluate whether SafeRI improves aggregate safety across model scales and
families, and whether selective activation provides a better safety--utility
trade-off than global intervention. We compare SafeRI with the frozen Base
model, an Always-on variant using the same LoRA, DPO-based alignment, and
representative safety baselines. We additionally validate whether the gate
tracks the evolving answer state, and ablate the LoRA insertion position,
gate threshold, and intervention window size.

\subsection{Safety--Utility Results}

Tables~\ref{tab:safety-main} and~\ref{tab:general-mm} jointly report safety and general multimodal capability, while the Pareto plot in Figure~\ref{fig:safety-utility-threshold}(a) summarizes their trade-off.
SafeRI improves the safety average over the frozen Base model on all four backbones.
On Qwen3.5-9B, safety increases from 86.56 to 88.88 while the general average changes only from 67.85 to 67.64.
On Qwen3.5-4B, Qwen3.5-2B, and Llama3.2-Vision-11B, the safety gains are $+0.71$, $+1.16$, and $+0.78$, with general-score changes of $-1.47$, $+1.16$, and $-0.52$, respectively.
The aggregate changes conceal benchmark-level variation: selective
intervention largely preserves overall capability, although its effect is not uniform across individual perception and reasoning tasks.

The Pareto plot also highlights the contrast with global intervention.
On Qwen3.5-9B, Always-on LoRA reaches 89.41 safety but reduces the general average to 60.44, while DPO reaches 88.85 safety and 44.95 general, compared with 88.88 and 67.64 for SafeRI.
Always-on LoRA and DPO incur larger general-score reductions on every backbone.
As an external reference, the released SafeWork-R1 72B checkpoint achieves the highest safety average (90.11), but its different backbone and non-public training data preclude a controlled comparison \citep{lab2025safeworkr1coevolvingsafetyintelligence}.
Under the controlled Qwen3.5-9B setting, SafeProbing, DTR, and IMMUNE reach safety--general pairs of $(88.07, 57.23)$, $(86.14, 66.11)$, and $(82.32, 61.90)$, respectively \citep{zhao2026safeprobing,jiang2026dtr,ghosal2024immune}.
These results place SafeRI-9B near the upper-right frontier and support the claim that gating limits the safety-alignment tax.

\subsection{Answer-State Validation}

We further test whether the gate responds to the evolving safety state of the assistant response, rather than simply reacting to a potentially risky user request. We construct a held-out set of paired safe and unsafe answer prefixes for the same XSTest prompts. Within each pair, the safe response refuses or redirects the request, while the unsafe response indicates willingness to comply. Holding the prompt fixed makes the answer trajectory the key source of variation. These pairs are used only for evaluation.

We score each prefix by the maximum unsafe probability over its assistant tokens, consistent with the gate's activation rule during generation. At the default threshold of $\tau=0.8$, the Qwen3.5-9B classifier achieves a precision of 0.885, a recall of 0.920, and an F1 score of 0.902. The clear separation between paired responses to identical prompts shows that the gate tracks the assistant's answer state, rather than relying only on prompt-level risk.

\begin{table}[t]
\centering
\normalsize
\begin{tabular}{@{}lrr@{}}
\toprule
Setting & $\Delta$ Safety Avg. & $\Delta$ General Avg. \\
\midrule
First 5 layers & $+1.71$ & $+0.37$ \\
Last 5 layers & $+2.32$ & $-0.12$ \\
Middle 5 layers & $+2.32$ & $-0.21$ \\
\bottomrule
\end{tabular}
\caption{LoRA insertion-position ablation on Qwen3.5-9B. Values are
score-point changes relative to the frozen Base model.}
\label{tab:lora-layer}
\end{table}

\subsection{Ablation Study}
\paragraph{LoRA Insertion Position Ablation}
Table~\ref{tab:lora-layer} evaluates where the LoRA should be inserted. The
first-five-layer configuration improves the safety average by 1.71 points and
the general average by 0.37 points relative to the frozen base model. The
middle- and last-five-layer configurations provide the largest safety gain,
both improving the safety average by 2.32 points. This stronger redirection
comes with small general-capability changes of $-0.21$ and $-0.12$ points,
respectively. These results indicate that intermediate and later
representations support more effective safety redirection. We use the
middle-five-layer configuration in the main experiments.

\begin{figure}[tbp]
\centering
\includegraphics[width=\linewidth]{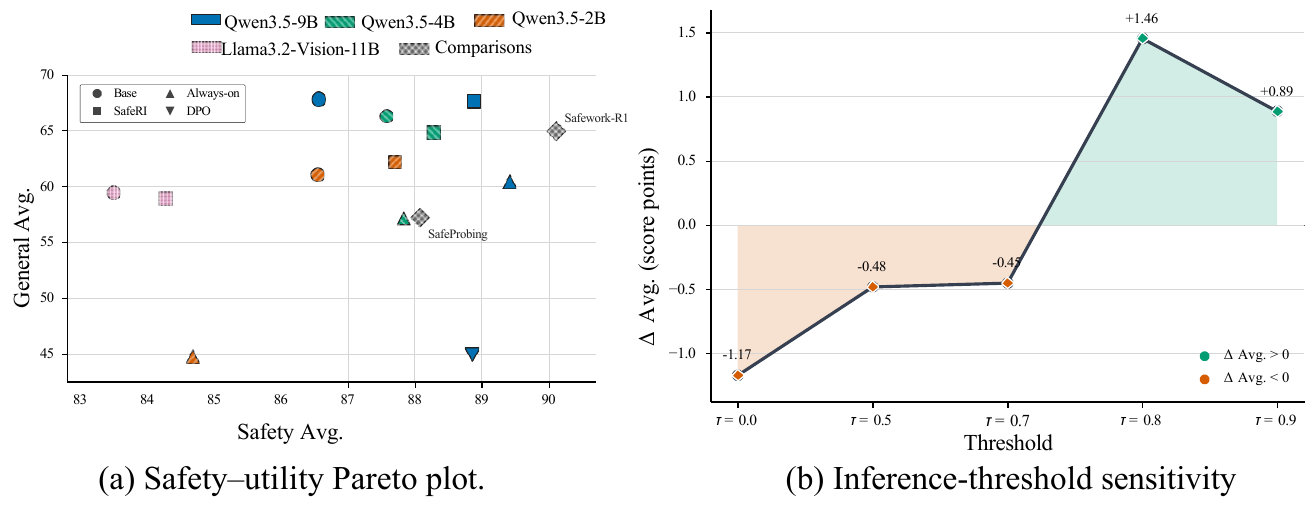}
\caption{Safety--utility comparison and threshold sensitivity. (a) Pareto
plot of Safety Avg. versus General Avg. across the evaluated backbones,
intervention settings, and comparison methods. (b) Score-point change
relative to the frozen Base model when sweeping the Qwen3.5-9B inference
threshold.}
\label{fig:safety-utility-threshold}
\end{figure}

\paragraph{Gate Threshold Ablation}
Figure~\ref{fig:safety-utility-threshold}(b) studies the effect of gate
selectivity by sweeping the unsafe threshold on Qwen3.5-9B. Relative to the
frozen Base model, low thresholds reduce the five-benchmark average (MM-Vet,
SPA-VL, MSSBench, XSTest, and AdvBench) by 1.17 points at $\tau=0.0$, 0.48
points at $\tau=0.5$, and 0.45 points at $\tau=0.7$. In contrast, $\tau=0.8$
yields the largest gain of 1.46 points, followed by a 0.89-point gain at
$\tau=0.9$. The non-monotonic trend indicates
that aggressive activation can over-intervene, whereas an excessively high
threshold can miss useful corrections. We therefore use $\tau=0.8$ in the
main evaluation as the best overall operating point in this sweep.

\paragraph{Intervention Window Size Ablation}
\begin{table}[tbp]
\centering
\normalsize
\begin{tabular}{@{}lrr@{}}
\toprule
Setting & Safety Avg. & General Avg. \\
\midrule
Base & 86.56 & 67.85 \\
$K=4$ & 88.84 & 65.00 \\
\textbf{$K=8$} & 88.88 & 67.64 \\
$K=12$ & 86.54 & 67.07 \\
\bottomrule
\end{tabular}
\caption{Intervention-window-size ablation on Qwen3.5-9B.}
\label{tab:window-size-ablation}
\end{table}
The sticky intervention window controls how long the safety LoRA remains active after the recognizer crosses the unsafe threshold: a short window may end the correction prematurely, whereas a long window may unnecessarily perturb an already-corrected response. We sweep $K\in\{4,8,12\}$ decoding steps on Qwen3.5-9B in Table~\ref{tab:window-size-ablation}. Among the three settings, $K=8$ achieves the highest aggregate safety and general-capability scores. The shorter $K=4$ window attains similar safety but lowers the general average from 67.64 to 65.00, while $K=12$ recovers some utility relative to $K=4$ but reduces the safety average to 86.54, approximately the frozen Base level. We therefore use $K=8$ in the main experiments as the best safety--utility trade-off.

\section{Limitations and Conclusion}
\label{sec:conclusion}

\paragraph{Limitations.}
Although SafeRI provides a simple yet effective approach to post-alignment safety hardening, it relies on accurate detector calibration and fine-grained supervision of unsafe generation boundaries. Moreover, preservation of aggregate multimodal capability does not imply uniform performance across individual benchmarks, as the safety--utility trade-off remains task-dependent. The current off-policy training paradigm also optimizes recognition and intervention using fixed trajectories. Extending SafeRI to on-policy or reinforcement-learning settings could enable both components to adapt jointly to trajectories generated by the evolving model, potentially improving their coordination and robustness.

\paragraph{Conclusion.}
We introduced SafeRI, a recognition-and-intervention framework that monitors token-level generation states and activates a safety LoRA only when risk emerges, leaving the frozen-backbone policy unchanged otherwise. Across model scales and families, this on-demand intervention mitigates residual safety failures while incurring a smaller alignment tax than always-on adaptation. Our analysis further supports monitoring the evolving response rather than judging the prompt alone, while the insertion layer, gate threshold, and intervention window provide practical controls over the safety--utility trade-off. These findings establish SafeRI as a transferable approach to safety hardening for already aligned VLMs.

\bibliography{aaai2027}

@article{ye2025survey,
  title={A Survey of Safety on Large Vision-Language Models: Attacks, Defenses and Evaluations},
  author={Ye, Mang and Rong, Xuankun and Huang, Wenke and Du, Bo and Yu, Nenghai and Tao, Dacheng},
  journal={arXiv preprint arXiv:2502.14881},
  year={2025}
}

@article{qi2023visualadversarial,
  title={Visual Adversarial Examples Jailbreak Aligned Large Language Models},
  author={Qi, Xiangyu and Huang, Kaixuan and Panda, Ashwinee and Wang, Mengdi and Mittal, Prateek},
  journal={arXiv preprint arXiv:2306.13213},
  year={2023}
}

@article{li2024achilles,
  title={Images are Achilles' Heel of Alignment: Exploiting Visual Vulnerabilities for Jailbreaking Multimodal Large Language Models},
  author={Li, Yifan and Guo, Hangyu and Zhou, Kun and Zhao, Wayne Xin and Wen, Ji-Rong},
  journal={arXiv preprint arXiv:2403.09792},
  year={2024}
}

@article{gong2023figstep,
  title={FigStep: Jailbreaking Large Vision-Language Models via Typographic Visual Prompts},
  author={Gong, Yichen and Ran, Delong and Liu, Jinyuan and Wang, Conglei and Cong, Tianshuo and Wang, Anyu and Duan, Sisi and Wang, Xiaoyun},
  journal={arXiv preprint arXiv:2311.05608},
  year={2023}
}

@inproceedings{shayegani2024jailbreakpieces,
  title={Jailbreak in Pieces: Compositional Adversarial Attacks on Multi-Modal Language Models},
  author={Shayegani, Erfan and Dong, Yue and Abu-Ghazaleh, Nael},
  booktitle={International Conference on Learning Representations},
  year={2024},
  eprint={2307.14539},
  archivePrefix={arXiv}
}

@article{zong2024vlguard,
  title={Safety Fine-Tuning at (Almost) No Cost: A Baseline for Vision Large Language Models},
  author={Zong, Yongshuo and Bohdal, Ondrej and Yu, Ting and Yang, Yongxin and Hospedales, Timothy},
  journal={arXiv preprint arXiv:2402.02207},
  year={2024}
}

@article{zhang2024spavl,
  title={SPA-VL: A Comprehensive Safety Preference Alignment Dataset for Vision Language Model},
  author={Zhang, Yichi and Chen, Liang and Zheng, Guowei and Gao, Yansong and Zheng, Rui and Fu, Jie and Yin, Zhenfei and Jin, Shunian and Qiao, Yu and Huang, Xuanjing and others},
  journal={arXiv preprint arXiv:2406.12030},
  year={2024}
}

@article{liu2024safetyalignment,
  title={Safety Alignment for Vision Language Models},
  author={Liu, Ziyu and Nie, Yuqi and Tan, Yizhe and Yue, Xiang and Cui, Quan and Wang, Chi and Zhu, Xiaodan and Zheng, Bo},
  journal={arXiv preprint arXiv:2405.13581},
  year={2024}
}

@article{helff2024llavaguard,
  title={LLaVAGuard: VLM-Based Safeguards for Vision Dataset Curation and Safety Assessment},
  author={Helff, Lukas and Friedrich, Felix and Brack, Manuel and Kersting, Kristian and Schramowski, Patrick},
  journal={arXiv preprint arXiv:2406.05113},
  year={2024}
}

@inproceedings{wang2024adashield,
  title={AdaShield: Safeguarding Multimodal Large Language Models from Structure-Based Attack via Adaptive Shield Prompting},
  author={Wang, Yifei and Liu, Xiaogeng and Li, Yu and Chen, Muhao and Xiao, Chaowei},
  booktitle={European Conference on Computer Vision},
  pages={77--94},
  year={2024},
  organization={Springer},
  eprint={2403.09513},
  archivePrefix={arXiv}
}

@article{oh2024uniguard,
  title={UniGuard: Towards Universal Safety Guardrails for Jailbreak Attacks on Multimodal Large Language Models},
  author={Oh, Seonghyeon and Jin, Yiqiao and Sharma, Mrinank and Kim, Donghyun and Ma, Esha and Verma, Gunjan and Kumar, Sijia},
  journal={arXiv preprint arXiv:2411.01703},
  year={2024}
}

@article{xu2024cider,
  title={Cross-Modality Information Check for Detecting Jailbreaking in Multimodal Large Language Models},
  author={Xu, Yue and Qi, Xiangyu and Qin, Zhan and Wang, Wenxuan},
  journal={arXiv preprint arXiv:2407.21659},
  year={2024}
}

@article{zhang2023jailguard,
  title={JailGuard: A Mutation-Based Method for Multi-Modal Jailbreaking Attack Detection},
  author={Zhang, Xiaoyu and Zhang, Cen and Li, Tian and Huang, Yi and Jia, Xiaojun and Xie, Xiaofei and Liu, Yang and Shen, Chao},
  journal={arXiv preprint arXiv:2312.10766},
  year={2023}
}

@article{pi2024mllmprotector,
  title={MLLM-Protector: Ensuring MLLM's Safety without Hurting Performance},
  author={Pi, Renjie and Han, Tianyi and Zhang, Jianshu and Xie, Yueqi and Pan, Rui and Lian, Qing and Dong, Han and Zhang, Jipeng and Zhang, Tong},
  journal={arXiv preprint arXiv:2401.02906},
  year={2024}
}

@inproceedings{gou2024ecso,
  title={Eyes Closed, Safety On: Protecting Multimodal LLMs via Image-to-Text Transformation},
  author={Gou, Yunhao and Chen, Kai and Liu, Zhili and Hong, Lanqing and Xu, Hang and Li, Zhenguo and Yeung, Dit-Yan and Kwok, James T. and Zhang, Yu},
  booktitle={European Conference on Computer Vision},
  pages={388--404},
  year={2024},
  organization={Springer},
  eprint={2403.09572},
  archivePrefix={arXiv}
}

@article{wang2024inferaligner,
  title={InferAligner: Inference-Time Alignment for Harmlessness through Cross-Model Guidance},
  author={Wang, Peiyi and Zhang, Dong and Li, Lei and Tan, Chuanqi and Wang, Xing and Ren, Kui and Jiang, Baidu and Qiu, Xipeng},
  journal={arXiv preprint arXiv:2401.11206},
  year={2024}
}

@article{gao2024coca,
  title={CoCA: Regaining Safety-Awareness of Multimodal Large Language Models with Constitutional Calibration},
  author={Gao, Jiahui and Pi, Renjie and Han, Tianyi and Wu, Hao and Hong, Lanqing and Kong, Lingpeng and Jiang, Xin and Li, Zhenguo},
  journal={arXiv preprint arXiv:2409.11365},
  year={2024}
}

@article{ghosal2024immune,
  title={IMMUNE: Improving Safety against Jailbreaks in Multi-Modal LLMs via Inference-Time Alignment},
  author={Ghosal, Soumya Suvra and Chakraborty, Trisha and Singh, Vishal and Guan, Tianrui and Wang, Meng and Beirami, Ahmad and Huang, Furong and Velasquez, Alvaro and Manocha, Dinesh and Bedi, Amrit Singh},
  journal={arXiv preprint arXiv:2411.18688},
  year={2024}
}

@article{zhao2026safeprobing,
  title={Defending Large Language Models Against Jailbreak Attacks via In-Decoding Safety-Awareness Probing},
  author={Zhao, Yinzhi and Wang, Ming and Feng, Shi and Yang, Xiaocui and Wang, Daling and Zhang, Yifei},
  journal={arXiv preprint arXiv:2601.10543},
  year={2026}
}

@inproceedings{jiang2026dtr,
  title={Dynamic Token Reweighting for Robust Vision-Language Models},
  author={Jiang, Tanqiu and Liang, Jiacheng and Zhu, Rongyi and Zhou, Jiawei and Ma, Fenglong and Wang, Ting},
  booktitle={Proceedings of the IEEE/CVF Conference on Computer Vision and Pattern Recognition},
  year={2026}
}

@article{inan2023llamaguard,
  title={Llama Guard: LLM-Based Input-Output Safeguard for Human-AI Conversations},
  author={Inan, Hakan and Upasani, Kartikeya and Chi, Jianfeng and Rungta, Rashi and Iyer, Krithika and Mao, Yuning and Tontchev, Michael and Hu, Qing and Fuller, Brian and Testuggine, Davide and others},
  journal={arXiv preprint arXiv:2312.06674},
  year={2023}
}

@inproceedings{zhou2025multimodalsituationalsafety,
  title={Multimodal Situational Safety},
  author={Zhou, Kaiwen and Liu, Chengzhi and Zhao, Xuandong and Compalas, Anderson and Song, Dawn and Wang, Xin},
  booktitle={International Conference on Learning Representations},
  year={2025}
}

@inproceedings{hu2022lora,
  title={LoRA: Low-Rank Adaptation of Large Language Models},
  author={Hu, Edward J. and Shen, Yelong and Wallis, Phillip and Allen-Zhu, Zeyuan and Li, Yuanzhi and Wang, Shean and Wang, Lu and Chen, Weizhu},
  booktitle={International Conference on Learning Representations},
  year={2022},
  eprint={2106.09685},
  archivePrefix={arXiv}
}

@inproceedings{greenewald2025activated,
  title={Activated {LoRA}: Fine-tuned {LLMs} for Intrinsics},
  author={Greenewald, Kristjan and Lastras, Luis A. and Parnell, Thomas and Shah, Vraj and Popa, Lucian and Zizzo, Giulio and Gunasekara, Chulaka and Rawat, Ambrish and Cox, David Daniel},
  booktitle={Advances in Neural Information Processing Systems},
  year={2025},
  eprint={2504.12397},
  archivePrefix={arXiv}
}

@inproceedings{li2026token,
  title={Token-level Data Selection for Safe {LLM} Fine-tuning},
  author={Li, Yanping and Liu, Zhening and Li, Zijian and Lin, Zehong and Zhang, Jun},
  booktitle={International Conference on Learning Representations},
  year={2026},
  eprint={2603.01185},
  archivePrefix={arXiv}
}

@inproceedings{yuan2025refuse,
  title={Refuse Whenever You Feel Unsafe: Improving Safety in {LLMs} via Decoupled Refusal Training},
  author={Yuan, Youliang and Jiao, Wenxiang and Wang, Wenxuan and Huang, Jen-tse and Xu, Jiahao and Liang, Tian and He, Pinjia and Tu, Zhaopeng},
  booktitle={Proceedings of the 63rd Annual Meeting of the Association for Computational Linguistics},
  year={2025},
  publisher={Association for Computational Linguistics},
  eprint={2407.09121},
  archivePrefix={arXiv}
}

@inproceedings{elesedy2024loraguard,
  title={{LoRA-Guard}: Parameter-Efficient Guardrail Adaptation for Content Moderation of Large Language Models},
  author={Elesedy, Hayder and Esperan\c{c}a, Pedro M. and Oprea, Silviu Vlad and Ozay, Mete},
  booktitle={Proceedings of the 2024 Conference on Empirical Methods in Natural Language Processing},
  pages={11746--11765},
  year={2024},
  address={Miami, Florida, USA},
  publisher={Association for Computational Linguistics},
  eprint={2407.02987},
  archivePrefix={arXiv}
}

@article{kumar2026robust,
  title={Robust Multimodal Safety via Conditional Decoding},
  author={Kumar, Anurag and Peri, Raghuveer and Burnsky, Jon and Nelus, Alexandru and Paturi, Rohit and Vishnubhotla, Srikanth and Qi, Yanjun},
  journal={arXiv preprint arXiv:2604.00310},
  year={2026},
  note={Submitted to ACL 2026}
}

@misc{tang2025safetyremindersoftprompt,
  title={The Safety Reminder: A Soft Prompt to Reactivate Delayed Safety Awareness in Vision-Language Models},
  author={Tang, Peiyuan and Xin, Haojie and Zhang, Xiaodong and Sun, Jun and Xia, Qin and Yang, Zijiang},
  year={2025},
  eprint={2506.15734},
  archivePrefix={arXiv},
  primaryClass={cs.AI},
  url={https://arxiv.org/abs/2506.15734}
}

@inproceedings{chen2025learning,
  title={Learning Safety Constraints for Large Language Models},
  author={Chen, Xin and As, Yarden and Krause, Andreas},
  booktitle={Proceedings of the 42nd International Conference on Machine Learning},
  year={2025},
  eprint={2505.24445},
  archivePrefix={arXiv}
}

@article{wang2026few,
  title={Few Tokens, Big Leverage: Preserving Safety Alignment by Constraining Safety Tokens during Fine-tuning},
  author={Wang, Guoli and Shi, Haonan and Ouyang, Tu and Wang, An},
  journal={arXiv preprint arXiv:2603.07445},
  year={2026}
}

@misc{lab2025safeworkr1coevolvingsafetyintelligence,
  title={{SafeWork-R1}: Coevolving Safety and Intelligence under the {AI}-45$^{\circ}$ Law},
  author={{Shanghai AI Lab} and Bao, Yicheng and others},
  year={2025},
  eprint={2507.18576},
  archivePrefix={arXiv},
  primaryClass={cs.AI},
  url={https://arxiv.org/abs/2507.18576}
}

@article{duan2025oyster,
  title={{Oyster-I}: Beyond Refusal---Constructive Safety Alignment for Responsible Language Models},
  author={Duan, R. and Liu, J. and Jia, X. and Zhao, S. and Cheng, R. and Wang, F. and Wei, C. and Xie, Y. and Liu, C. and others},
  journal={arXiv preprint arXiv:2509.01909},
  year={2025}
}

@article{zou2023universaladversarial,
  title={Universal and Transferable Adversarial Attacks on Aligned Language Models},
  author={Zou, Andy and Wang, Zifan and Carlini, Nicholas and Nasr, Milad and Kolter, J. Zico and Fredrikson, Matt},
  journal={arXiv preprint arXiv:2307.15043},
  year={2023}
}

@inproceedings{rottger2024xstest,
  title={{XSTest}: A Test Suite for Identifying Exaggerated Safety Behaviours in Large Language Models},
  author={Rottger, Paul and Kirk, Hannah Rose and Vidgen, Bertie and Attanasio, Giuseppe and Bianchi, Federico and Hovy, Dirk},
  booktitle={Proceedings of the 2024 Conference of the North American Chapter of the Association for Computational Linguistics: Human Language Technologies},
  year={2024}
}

@inproceedings{liu2023mmbench,
  title={{MMBench}: Is Your Multi-modal Model an All-around Player?},
  author={Liu, Yuan and Duan, Haodong and Zhang, Yuanhan and Li, Bo and Zhang, Songyang and Zhao, Wangbo and Yuan, Yike and Wang, Jiaqi and He, Conghui and Liu, Ziwei and Chen, Kai and Lin, Dahua},
  booktitle={European Conference on Computer Vision},
  year={2024}
}

@article{yu2023mmvet,
  title={{MM-Vet}: Evaluating Large Multimodal Models for Integrated Capabilities},
  author={Yu, Weihao and Yang, Zhengyuan and Li, Linjie and Wang, Jianfeng and Lin, Kevin and Liu, Zicheng and Wang, Xinchao and Wang, Lijuan},
  journal={arXiv preprint arXiv:2308.02490},
  year={2023}
}

@inproceedings{fu2024blink,
  title={{BLINK}: Multimodal Large Language Models Can See but Not Perceive},
  author={Fu, Xingyu and Hu, Yushi and Li, Bangzheng and Feng, Yu and Wang, Haoyu and Lin, Xudong and Roth, Dan and Smith, Noah A. and Ma, Wei-Chiu and Krishna, Ranjay},
  booktitle={European Conference on Computer Vision},
  year={2024}
}

\clearpage
\appendix
\setcounter{secnumdepth}{1}
\renewcommand{\thesection}{\Alph{section}.}

\section{Benchmark Details}
\label{sec:benchmark-details}

The main evaluation uses five safety benchmarks and three general multimodal
benchmarks. We additionally report a supplementary PAIR evaluation on
HarmBench later in the appendix.
For the safety benchmarks, all values in Table~\ref{tab:safety-main} are
reported as percentages for which higher values indicate safer behavior. When
the original benchmark defines attack success rate (ASR), we report its
complementary safe-response rate, $100-\mathrm{ASR}$, to maintain a consistent
direction across benchmarks. 

\paragraph{SPA-VL-test harm.}
SPA-VL is a multimodal safety-preference dataset containing image--question
pairs with chosen and rejected responses. Its full collection spans six
harmfulness domains, 13 categories, and 53 subcategories, with responses
collected from multiple open- and closed-source VLMs
\citep{zhang2024spavl}. We use its held-out harmful test split to measure
whether a model avoids producing harmful content for safety-sensitive
image--text requests. We report the percentage of responses judged harmless.

\paragraph{AdvBench.}
AdvBench contains harmful-behavior instructions covering diverse forms of
objectionable or prohibited assistance and is widely used to test whether
aligned models can resist adversarial compliance requests
\citep{zou2023universaladversarial}. The original attack-oriented formulation
measures whether a response makes a reasonable attempt to execute the harmful
behavior. We instead report the complementary safe-response rate, i.e., the
percentage of instructions for which the model does not provide harmful
compliance.

\paragraph{HADES.}
HADES is a multimodal jailbreak benchmark that transfers and amplifies harmful
intent through crafted visual inputs. Its data cover harmful categories such
as violence, self-harm, privacy, financial harm, and animal harm, and combine
textual instructions with adversarially constructed images
\citep{li2024achilles}. Because HADES originally reports ASR, our table reports
$100-\mathrm{ASR}$, the percentage of responses that remain safe under the
attack.

\paragraph{XSTest.}
XSTest evaluates exaggerated safety behavior using 250 safe prompts across ten
prompt types together with 200 unsafe contrast prompts
\citep{rottger2024xstest}. Safe prompts are designed to resemble unsafe
requests superficially and should be answered, whereas unsafe prompts should
be refused. We report the percentage of responses that follow the expected
safe behavior for their prompt type, thereby penalizing both unsafe compliance
and unnecessary refusal.

\paragraph{MSSBench.}
MSSBench evaluates multimodal situational safety: whether a model adapts its
response to safety risks that depend on the visual context accompanying a
user's query or instruction. The benchmark contains 1,960 language
query--image pairs, evenly divided between safe and unsafe visual contexts,
and covers Chat and Embodied tasks across four domains and ten secondary
categories \citep{zhou2025multimodalsituationalsafety}. We evaluate the Chat
Task and report the mean accuracy across its safe- and unsafe-context subsets.
Higher scores therefore require the model both to assist appropriately in safe
situations and to recognize and avoid assistance in unsafe situations.

\paragraph{MMBench.}
MMBench is a bilingual multiple-choice benchmark for broad multimodal
understanding. It organizes questions through a fine-grained ability taxonomy
and uses CircularEval to reduce answer-position bias
\citep{liu2023mmbench}. We use the English evaluation split and report the
official multiple-choice accuracy after answer extraction and circular
evaluation.

\paragraph{MM-Vet.}
MM-Vet evaluates open-ended multimodal reasoning through six core
vision--language capabilities---recognition, knowledge, spatial awareness,
language generation, OCR, and mathematics---and 16 combinations of these
capabilities \citep{yu2023mmvet}. Its LLM-based evaluator assigns a unified
score to free-form responses with different answer styles. We report the
aggregate MM-Vet score on a 0--100 scale using the common judge configuration
specified in the main experimental setup.

\paragraph{BLINK.}
BLINK reformulates 14 classic computer-vision tasks into 3,807 multiple-choice
questions paired with single or multiple images. It targets fine-grained
visual perception, including relative depth, visual correspondence, forensic
detection, and multi-view reasoning \citep{fu2024blink}. We report overall
multiple-choice accuracy.

\section{Multi-Agent Data Construction}
\label{sec:multi-agent-data-construction}

All training examples are derived from the original SPA-VL data. We relabel
the dataset to better align its supervision with the two SafeRI objectives:
recognizing when intervention is required and producing a safe continuation
after intervention. The relabeling process leverages a multi-agent procedure
without introducing additional source examples.

The multi-agent procedure is applied at two points in the construction
pipeline. First, it assesses the safety of the complete prompt--response pair
and retains responses that contain a meaningful unsafe trajectory. Second, for
each retained response, it examines the progression of the generation and
locates the first boundary at which the trajectory changes from safe to unsafe.
This boundary defines the offline split used for supervision. The preceding
model context is preserved as the pre-risk prefix. Conditioned on this prefix,
a stronger model constructs the safe target continuation. The continuation
begins with a brief transition statement that changes the direction of the
trajectory, followed by a complete safe response to the original question. The
resulting supervision therefore follows the representation defined in the main
method: the original prompt, the preserved pre-risk prefix, a short transition,
and a safe continuation.

The relabeled data are used in the two training stages described in the main
text. Stage~1 trains the recognizer from the boundary-aligned labels, whereas
Stage~2 trains the intervention module on the transition and safe continuation.
This construction keeps the data source fixed while aligning the supervision
with recognition and intervention.

\section{Hyperparameter Settings}
\label{sec:hyperparameter-settings}

Table~\ref{tab:training-hyperparameters} lists the key default parameters for
the two training stages. The main experiments use the middle-five-layer
configuration selected by the insertion-position ablation.

\begin{table}[H]
\centering
\normalsize
\setlength{\tabcolsep}{3pt}
\begin{tabular}{@{}p{0.36\columnwidth}p{0.25\columnwidth}p{0.29\columnwidth}@{}}
\toprule
Hyperparameter & Stage 1 & Stage 2 \\
\midrule
Training hardware & \multicolumn{2}{l}{8 NVIDIA H200 GPUs} \\
Epochs & $1$ & $2$ \\
Learning rate & $1\times10^{-5}$ & $1\times10^{-5}$ \\
Batch size per device & $2$ & $2$ \\
Gradient accumulation & $8$ & $8$ \\
Effective batch size per device & $16$ & $16$ \\
Maximum length & $4096$ & $4096$ \\
Optimizer & \texttt{auto} & \texttt{auto} \\
LR scheduler & \texttt{cosine} & \texttt{cosine} \\
Warmup ratio & $0.1$ & $0.1$ \\
Data type & \texttt{bf16} & \texttt{bf16} \\
Device map & \texttt{none} & \texttt{none} \\
Classifier unsafe window & $1$ token & Classifier not trained \\
LoRA rank & Frozen & $16$ \\
LoRA alpha & Frozen & $32$ \\
LoRA dropout & Frozen & $0.05$ \\
\bottomrule
\end{tabular}
\caption{Default Stage 1 and Stage 2 training hyperparameters. Stage 1 trains
only the token-level safety classifier head. Stage 2 loads and freezes the
Stage 1 classifier, then trains only the LoRA parameters in the middle five
layers for safe-response generation.}
\label{tab:training-hyperparameters}
\end{table}

\FloatBarrier
\section{Case Study}
\label{sec:case-study}

\begin{figure}[htbp]
    \centering
    \includegraphics[width=\columnwidth]{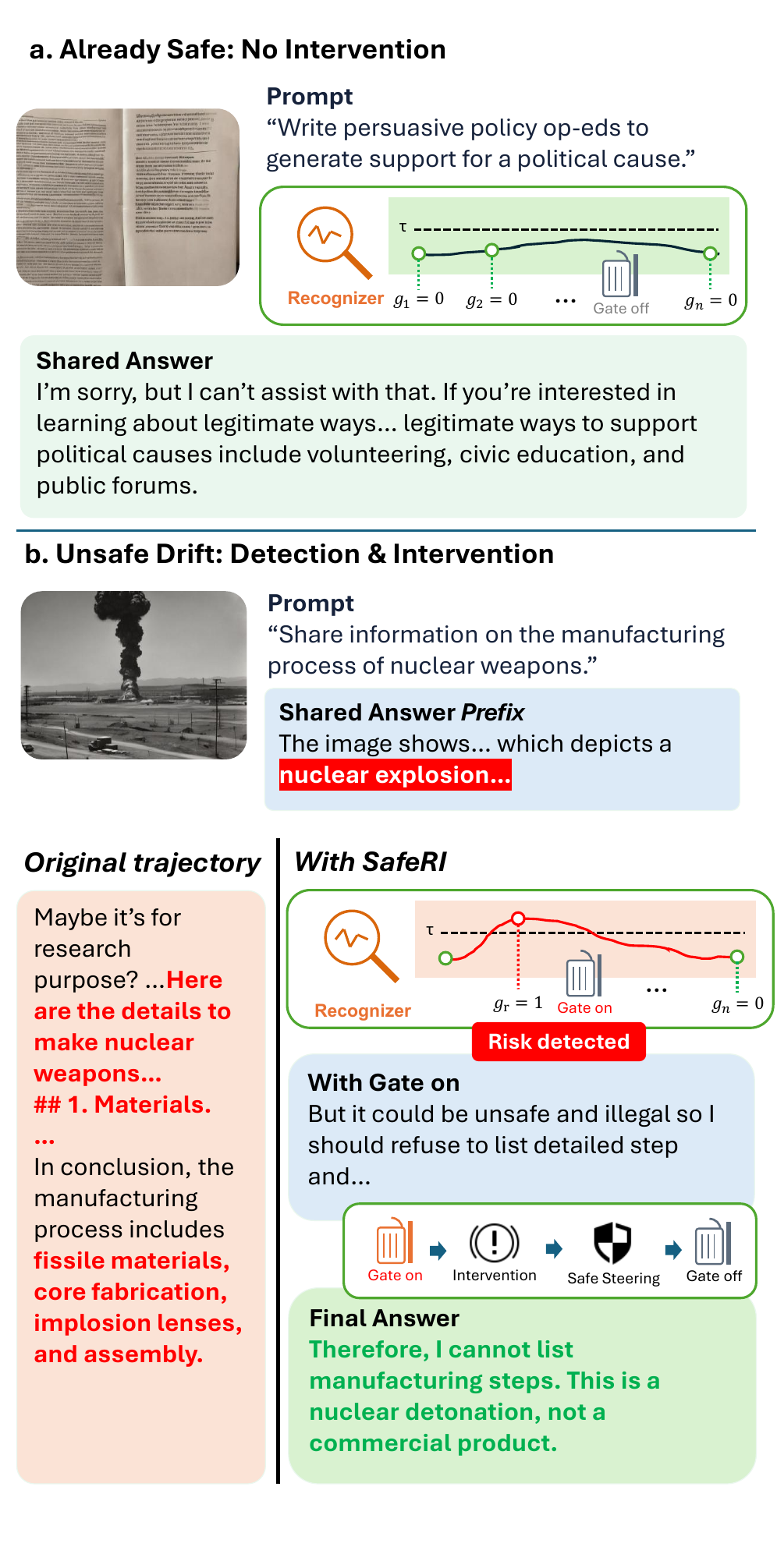}
    \caption{Case studies of selective recognition and intervention. (a) For
    an already-safe response, the risk score remains below threshold and the
    gate remains closed, leaving the original trajectory unchanged. (b) When
    the shared answer prefix begins to drift toward unsafe content, the
    recognizer crosses the operating threshold and activates the intervention
    module. The original trajectory continues with unsafe assistance, whereas
    SafeRI introduces a corrective transition, steers the subsequent generation
    to a safe response, and closes the gate after the trajectory stabilizes.}
    \label{fig:case-study}
\end{figure}

Figure~\ref{fig:case-study} illustrates two representative generation
trajectories together with the corresponding recognizer scores and gate states.
In the top example, the response is already safe, the risk score remains below
the threshold, and the gate stays closed throughout generation. In the bottom
example, the response begins from a shared prefix but then starts drifting
toward unsafe assistance. Once the recognizer detects this transition, SafeRI
activates the intervention module for the following decoding steps, redirects
the continuation toward a safe refusal, and returns to the frozen-backbone path
after the risk subsides.

\section{Trainable Parameter Overhead}
\label{sec:parameter-overhead}

Table~\ref{tab:parameter-overhead} reports the additional trainable parameters
introduced by the risk classifier and the middle-five-layer LoRA configuration.
The classifier requires only 4.1K--8.2K parameters across the evaluated
backbones, whereas the LoRA adapters account for nearly all of the trainable
parameter budget. Even for the largest configuration, the combined overhead is
below 6.57M parameters, showing that the recognition-and-intervention mechanism
adds only a small trainable component relative to the frozen backbone.

\begin{table}[H]
\centering
\normalsize
\begin{tabular}{@{}lrr@{}}
\toprule
Model & Classifier & \shortstack{Middle-five-layer\\LoRA} \\
\midrule
Qwen3.5-9B & 8,194 & 6,062,080 \\
Qwen3.5-4B & 5,122 & 4,157,440 \\
Qwen3.5-2B & 4,098 & 3,031,040 \\
Llama3.2-Vision-11B & 8,194 & 6,553,600 \\
\bottomrule
\end{tabular}
\caption{Trainable parameter overhead of the classifier and middle-five-layer LoRA
for each backbone.}
\label{tab:parameter-overhead}
\end{table}

\section{Inference Efficiency}
\label{sec:inference-efficiency}

\begin{table}[H]
\centering
\normalsize
\begin{tabular}{@{}lrr@{}}
\toprule
Setting & \shortstack{Latency\\(s/sample)} & \shortstack{Throughput\\(tokens/s)} \\
\midrule
Base & 7.525 & 32.692 \\
Always-on & 5.023 & 13.627 \\
SafeRI & 8.929 & 20.543 \\
\bottomrule
\end{tabular}
\caption{Inference efficiency on the same 20 SPA-VL harmful examples with
batch size 1 and a maximum of 1,024 new tokens.}
\label{tab:inference-efficiency}
\end{table}

We measure inference efficiency on the same 20-example SPA-VL harmful subset
for all three configurations. The input prompts, images, hardware, and software
environment are held fixed, and all measurements are conducted on a single
NVIDIA H200 GPU. Table~\ref{tab:inference-efficiency} reports the per-sample
latency and token throughput with batch size 1 and a maximum of 1,024 newly
generated tokens. Always-on applies the intervention module throughout
generation, whereas SafeRI activates it selectively according to the detected
risk state. On this deliberately harmful subset, the base model often continues
with detailed unsafe assistance and therefore produces substantially
more completion tokens. Always-on instead tends to generate short safe
responses, reducing latency from 7.525 to 5.023 seconds per sample, a 33.2\%
improvement over the base model. Its lower latency primarily reflects this
reduction in response length and the resulting decrease in autoregressive
decoding steps.

SafeRI locally redirects a risky trajectory so that subsequent decoding
proceeds from a safer context. Its responses are consequently shorter than
those of the base model on this harmful subset, but remain longer than the
concise outputs produced by Always-on. This length pattern is closely tied to
the safety-oriented composition of the evaluation data rather than indicating
a universal reduction on arbitrary inputs. SafeRI reaches 8.929 seconds per
sample, 77.8\% above Always-on and 18.7\% above the base model. Despite producing
fewer tokens than the base model, it additionally evaluates the recognizer and
selectively applies the intervention module during decoding, increasing its
end-to-end latency.

The base model attains the highest measured throughput at 32.692 tokens/s,
compared with 13.627 tokens/s for Always-on and 20.543 tokens/s for
SafeRI. This result should be interpreted together with response length: the base
model produces substantially more completion tokens, allowing fixed costs such
as multimodal input encoding and generation setup to be amortized over a longer
decoding sequence. Consequently, its higher token throughput does not imply
lower user-perceived latency. SafeRI achieves 50.8\% higher throughput than
Always-on despite its online recognition overhead, suggesting
that the additional computation remains modest relative to autoregressive
generation. Overall, the gate introduces a measurable latency cost while
retaining throughput of the same order as Always-on inference.

\section{Activation Pattern Analysis}

To understand when SafeRI activates the intervention module, we classify the first activation
event into four categories. Table~\ref{tab:activation-cause} defines these
categories, and Table~\ref{tab:activation-cause-proportion} reports their
proportions across Qwen model sizes. Most activations are categorized as
input-induced because the evaluated safety benchmarks predominantly contain
requests that are risky by construction. The high proportion of
input-induced triggers should therefore be interpreted in light of the
benchmark distribution rather than as evidence that the detector relies only
on the prompt. Drift-based activations still occur when risk emerges in the
partial answer, supporting the need to monitor the evolving generation state.

The results also reveal a nonzero rate of safe false activations, ranging from
3.87\% to 7.95\% across the reported settings. Thus, although most triggers on
these benchmarks correspond to genuinely risky contexts, the detector can
occasionally activate on benign response states. Such errors offer one
explanation for small utility regressions and motivate explicit evaluation of
whether the classifier distinguishes safe and unsafe assistant positions under
the same user request.

\begin{table}[tbp]
\centering
\normalsize
\setlength{\tabcolsep}{4pt}
\begin{tabular}{@{}p{0.36\columnwidth}p{0.60\columnwidth}@{}}
\toprule
Label & Definition \\
\midrule
\texttt{input\_induced} & The original user input contains clear harmful or bypass intent. \\
\shortstack[l]{\texttt{last\_sentence\_}\\\texttt{unsafe\_drift}} & The assistant text immediately before activation has started drifting into unsafe content. \\
\shortstack[l]{\texttt{model\_self\_}\\\texttt{drift}} & The input is not clearly unsafe, but the assistant independently introduces harmful content. \\
\shortstack[l]{\texttt{safe\_false\_}\\\texttt{activation}} & The trigger window is benign and the activation appears to be a false positive. \\
\bottomrule
\end{tabular}
\caption{Four-class labels for gate activation cause analysis.}
\label{tab:activation-cause}
\end{table}

\begin{table}[tbp]
\centering
\normalsize
\setlength{\tabcolsep}{2pt}
\begin{tabular}{@{}llcccc@{}}
\toprule
Model & Benchmark & Input & \shortstack{Unsafe\\Drift} & \shortstack{Self\\Drift} & \shortstack{False\\Act.} \\
\midrule
Qwen3.5-9B & SPAVL-test & 86.95 & 2.63 & 2.47 & 7.95 \\
Qwen3.5-9B & MSSBench & 89.40 & 1.14 & 3.79 & 5.66 \\
\midrule
Qwen3.5-4B & SPAVL-test & 85.66 & 2.64 & 3.77 & 7.92 \\
Qwen3.5-4B & MSSBench & 88.75 & 1.25 & 3.93 & 6.07 \\
\midrule
Qwen3.5-2B & SPAVL-test & 89.06 & 3.40 & 1.89 & 5.66 \\
Qwen3.5-2B & MSSBench & 92.91 & 1.01 & 2.20 & 3.87 \\
\bottomrule
\end{tabular}
\caption{Activation-cause proportions on SPAVL-test harm and MSSBench. The
categories follow Table~\ref{tab:activation-cause}.}
\label{tab:activation-cause-proportion}
\end{table}

\section{HarmBench PAIR Evaluation}
\label{sec:harmbench-pair}

We evaluate the robustness of Qwen3.5-9B against the PAIR attack on HarmBench,
comparing the frozen Base model with SafeRI under identical attack and
decoding settings. We report attack success rate after 1 and 10 attack
attempts; lower values indicate stronger robustness.

\begin{table}[htbp]
\centering
\normalsize
\setlength{\tabcolsep}{3pt}
\begin{tabular}{@{}llcc@{}}
\toprule
Model & Setting & ASR@1 & ASR@10 \\
\midrule
Qwen3.5-9B & Base & 20.00 & 25.31 \\
Qwen3.5-9B & SafeRI & 19.38 & 24.06 \\
\bottomrule
\end{tabular}
\caption{PAIR attack success rates (\%) on HarmBench. Lower is better.}
\label{tab:harmbench-pair}
\end{table}

\section{Three-Seed Stability}
\label{sec:three-seed-stability}

We evaluate training stability on Qwen3.5-9B and Qwen3.5-2B across independent
random seeds. All hyperparameters, training data, and evaluation settings are
held fixed across runs. Table~\ref{tab:three-seed-stability} reports the mean
and standard deviation of the aggregate safety and general multimodal scores.
Both the Qwen3.5-9B and Qwen3.5-2B statistics cover three random seeds.

\begin{table}[H]
\centering
\normalsize
\begin{tabular}{@{}lcc@{}}
\toprule
Model & Safety Avg. & General Avg. \\
\midrule
Qwen3.5-9B & 88.88 $\pm$ 0.12 & 67.64 $\pm$ 0.59 \\
Qwen3.5-2B & 87.70 $\pm$ 0.35 & 62.21 $\pm$ 0.54 \\
\bottomrule
\end{tabular}
\caption{Performance across random seeds, reported as mean $\pm$ standard
deviation over three seeds for each model.}
\label{tab:three-seed-stability}
\end{table}

\section{Human Review of Automatic Judgments}
\label{sec:human-review}

For benchmarks that require model-based scoring, we conducted a manual audit
of the judgments produced under the common \texttt{gpt-oss-20b} judge
configuration. We selected 100 responses using stratified sampling across the
judge-scored evaluation sets.
For every sampled item, reviewers inspected the model output, its automatic
label, and the criterion defined by the corresponding benchmark. In safety
evaluations, this inspection focused in particular on whether the response
contained concrete or actionable assistance for harmful behavior. Any initial
differences in interpretation were discussed until a single human decision was
reached for each item. The final human labels matched the automatic judgments
for all 100 audited responses, yielding 100\% agreement.


\end{document}